\documentclass{article}
\usepackage{spconf}

\usepackage[labelfont=bf]{caption}                         
\usepackage[subrefformat=parens]{subfig}
\usepackage{enumitem}  
\usepackage{amssymb}
\usepackage[misc]{ifsym}
\usepackage{graphicx}
\usepackage{threeparttable}
\usepackage{ntheorem}
\usepackage{color}
\usepackage{colortbl}
\usepackage{multirow}
\usepackage{amsmath}
\usepackage{comment}
\usepackage{bm}                                            
\usepackage{etoolbox}                                      
\usepackage{url}
\usepackage{nth}
\usepackage{cite}
\usepackage{balance}
\usepackage[bookmarks=false]{hyperref}                     %
\usepackage{courier}                                       
\usepackage{listings}                                      
\usepackage[]{algpseudocode}                               
\algrenewcommand\textproc{\texttt}
\makeatletter\let\c@float@type\relax\makeatother
\makeatletter\let\float@addtolists\relax\makeatother
\usepackage{algorithm}

\usepackage{amsmath,lipsum}

\title{Geometric Loss for Deep Multiple Sclerosis lesion Segmentation}
\name{
\parbox{\linewidth}{\centering
Hang Zhang$^{\star \dagger}$  ~ Jinwei Zhang$^{\star \dagger}$ ~  Rongguang Wang$^{\ddag}$ ~ Qihao Zhang$^{\star \dagger}$ ~ Susan A. Gauthier$^{\dagger}$ \\ Pascal Spincemaille$^{\dagger}$ ~ Thanh D. Nguyen$^{\dagger}$ ~ Yi Wang$^{\star \dagger}$
}
} 

\address{$^{\star}$ Cornell University, $^{\dagger}$ Weill Cornell Medical College, $^{\ddag}$ University of Pennsylvania}

\begin{document}
%
\maketitle

\begin{abstract}
Multiple sclerosis (MS) lesions occupy a small fraction of the brain volume, and are heterogeneous with regards to shape, size and locations, which poses a great challenge for training deep learning based segmentation models.
We proposed a new geometric loss formula to address the data imbalance and exploit the geometric property of MS lesions.
We showed that traditional region-based and boundary-aware loss functions can be associated with the formula.
We further develop and instantiate two loss functions containing first- and second-order geometric information of lesion regions to enforce regularization on optimizing deep segmentation models.
Experimental results on two MS lesion datasets with different scales, acquisition protocols and resolutions demonstrated the superiority of our proposed methods compared to other state-of-the-art methods. 
\end{abstract}

\begin{keywords}
Geometric Transformation, Data Imbalance, Image Segmentation, Multiple Sclerosis Lesion
\end{keywords}
\section{Introduction}

\begin{figure}[!t]
	\centering
	\subfloat[T1]{\includegraphics[width=.28\columnwidth]{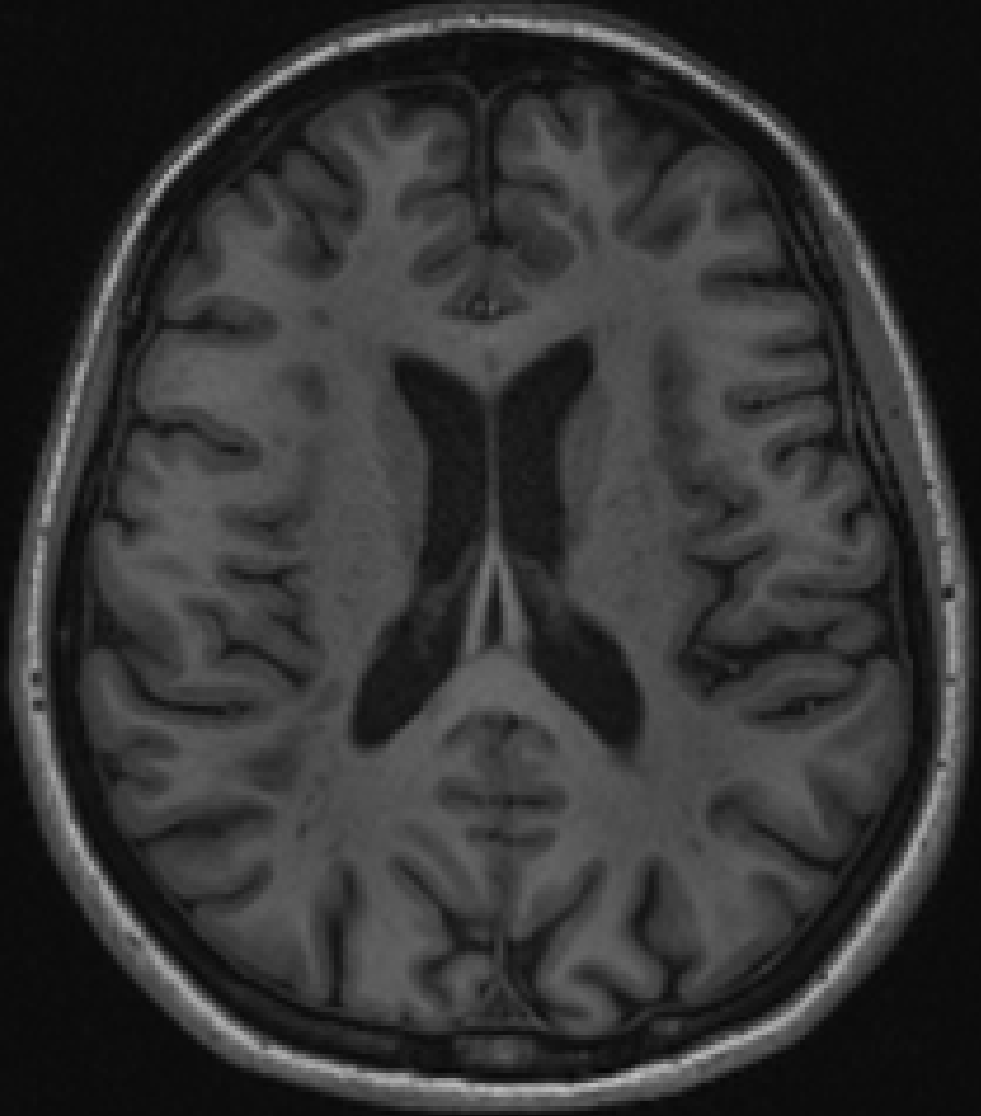}}
	\hspace{0.5ex}
	\subfloat[T2]{\includegraphics[width=.28\columnwidth]{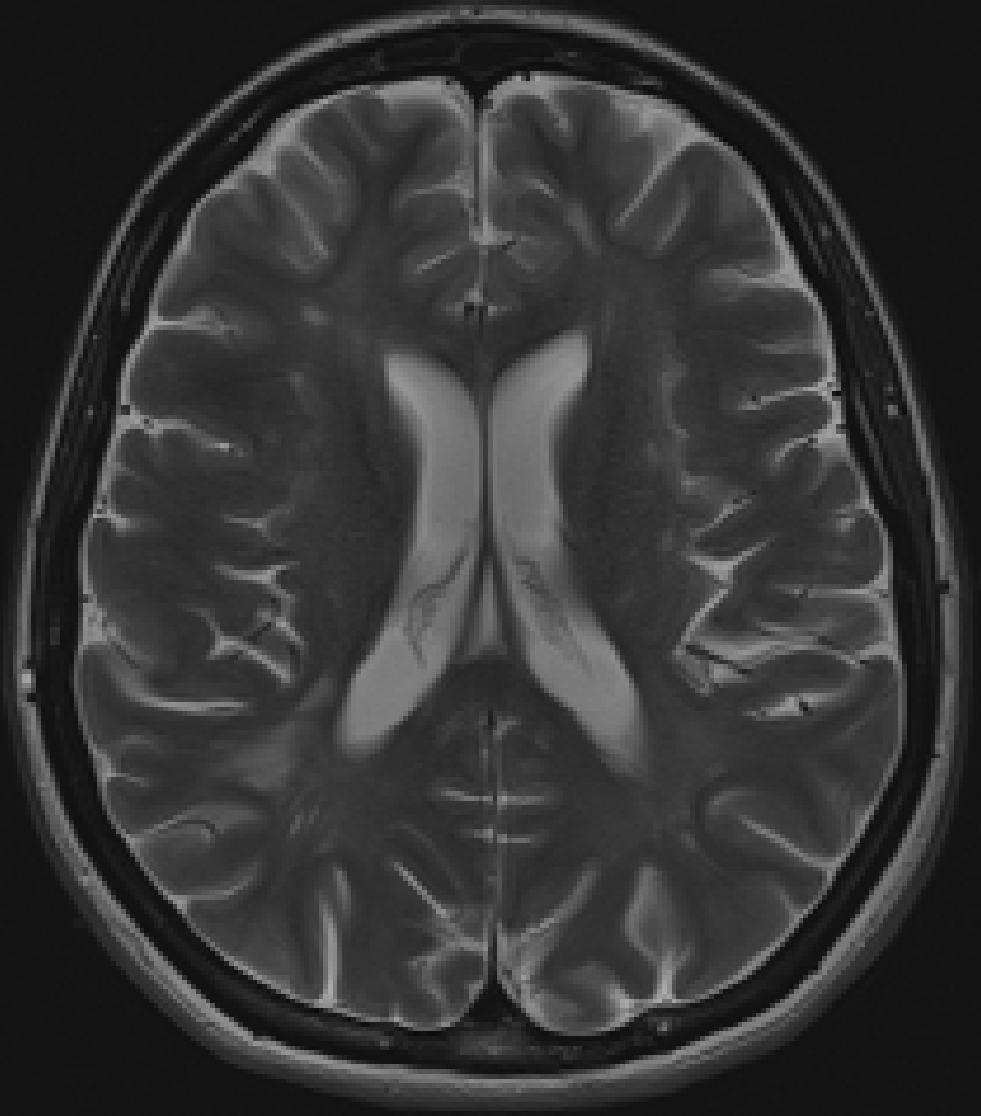}} 
	\hspace{0.5ex}
	\subfloat[T2-FLAIR]{\includegraphics[width=.28\columnwidth]{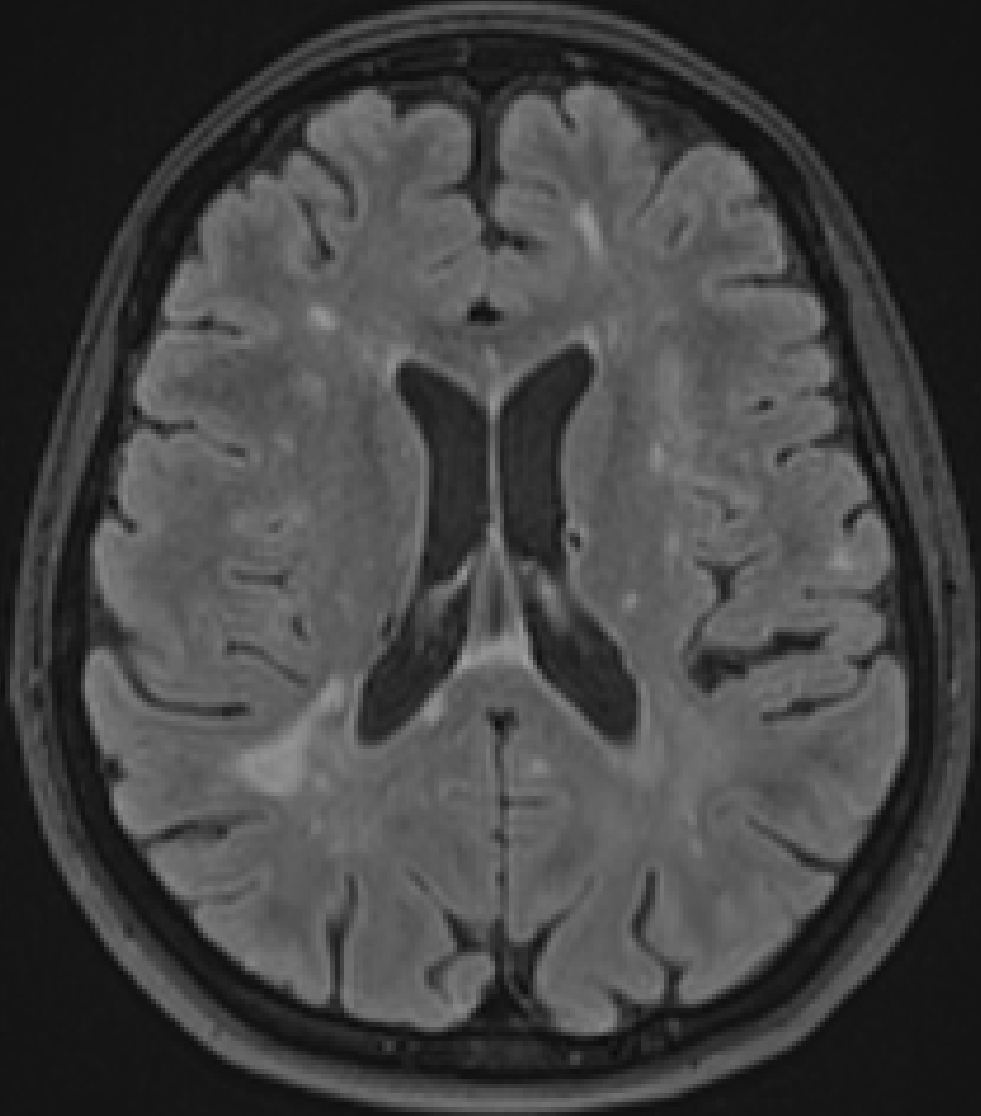}}
	\\
	\vspace{-1.5ex}
	\subfloat[Ground-truth]{\includegraphics[width=.28\columnwidth]{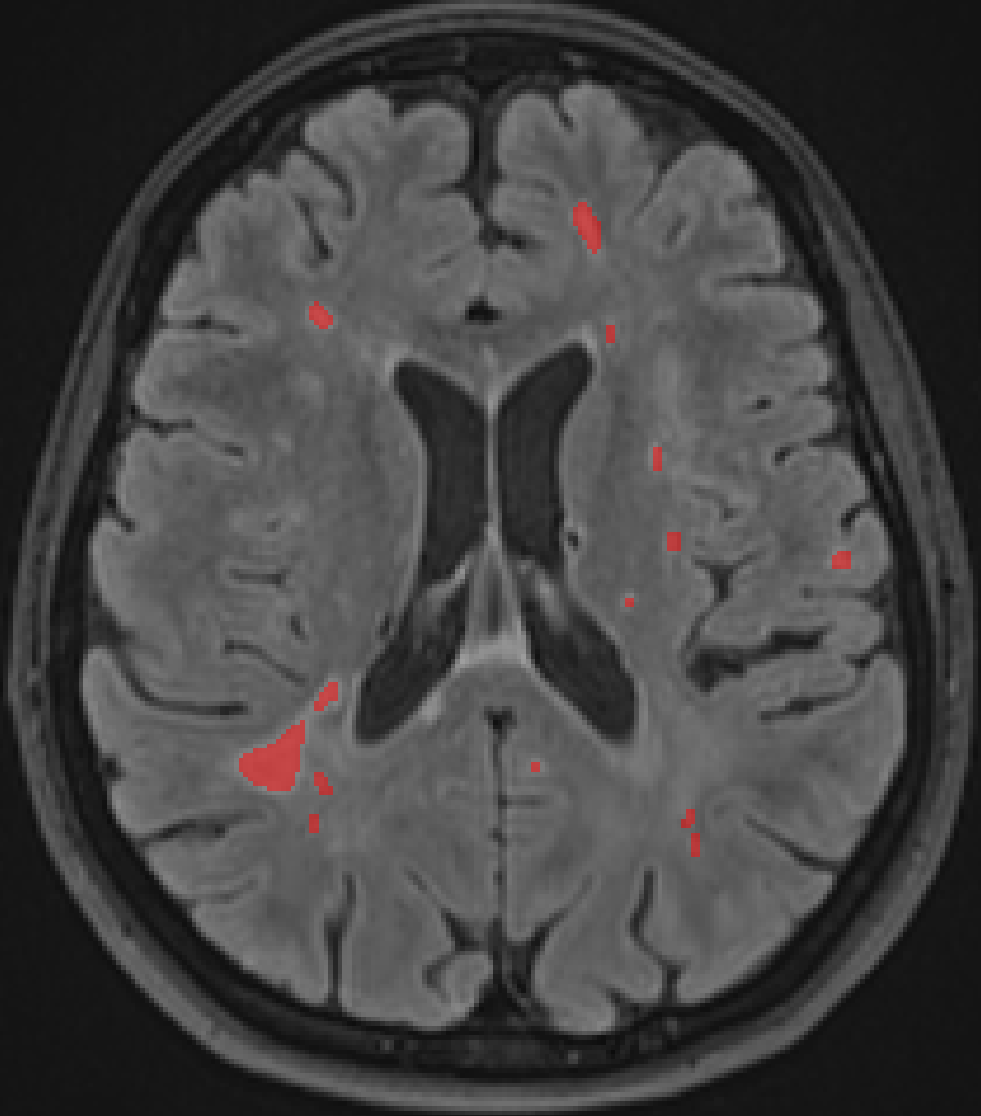}}
	\hspace{0.5ex}
	\subfloat[BD]{\includegraphics[width=.28\columnwidth]{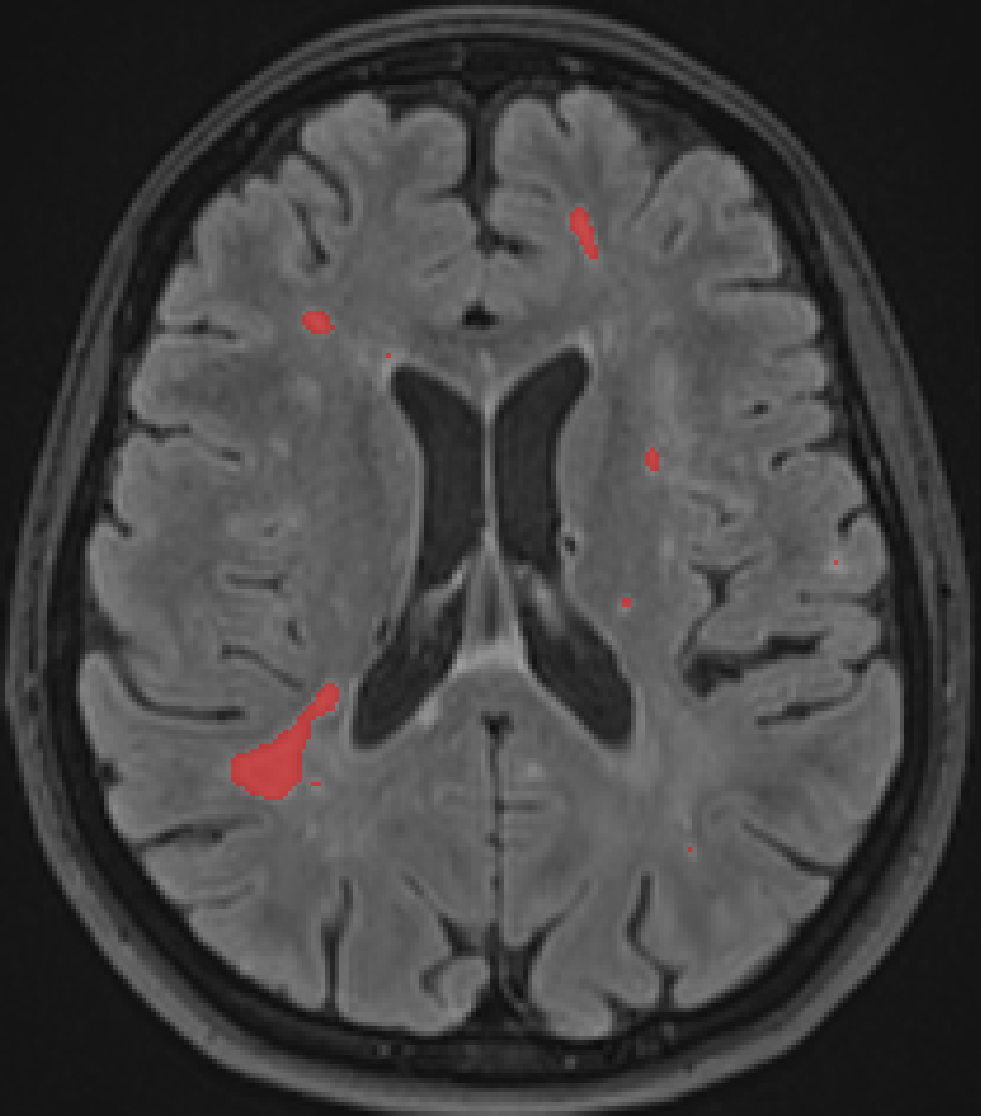}} 
    \hspace{0.5ex}
	\subfloat[FOG]{\includegraphics[width=.28\columnwidth]{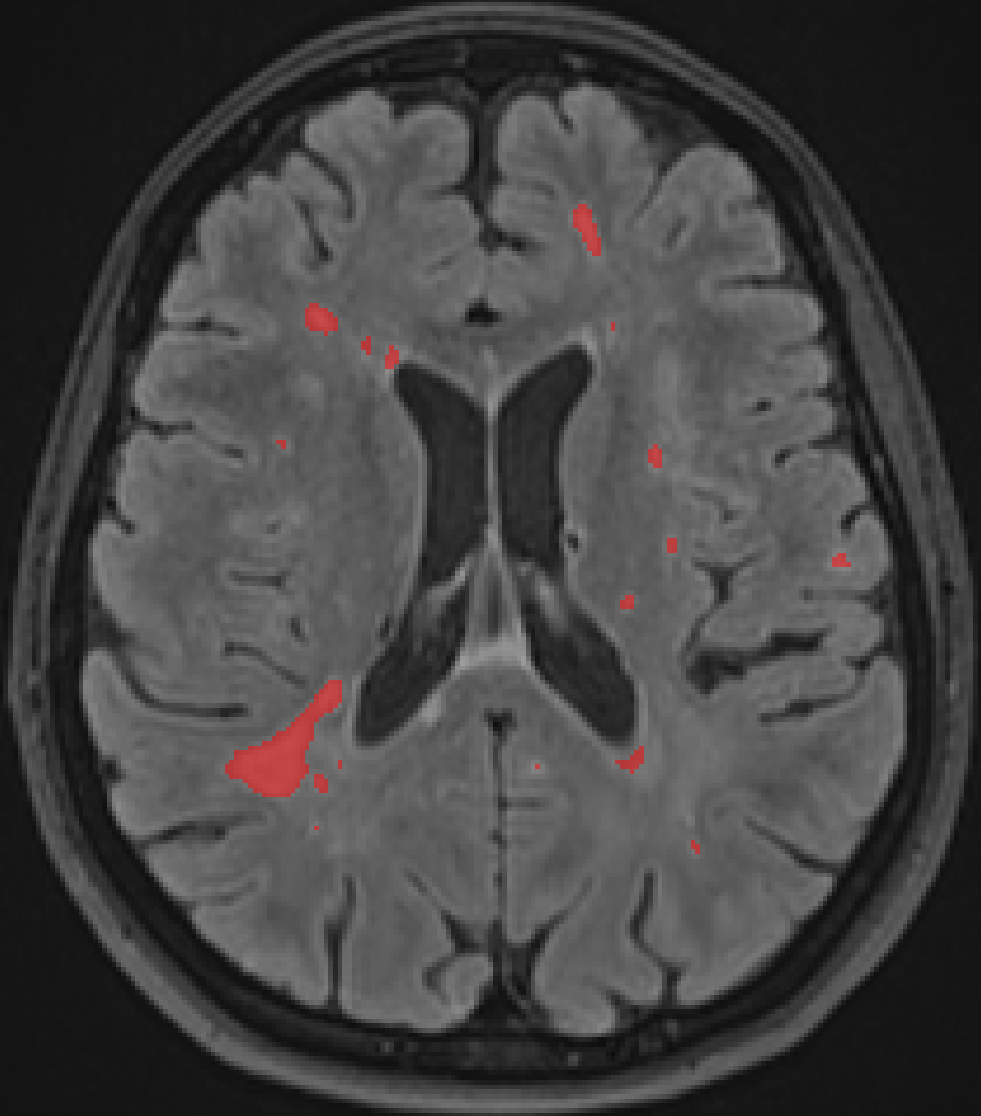}}
	\caption{ 
	    Example of T1, T2, T2-FLAIR images and corresponding lesion segmentation  masks produced by a human expert, and models trained by BD and FOG loss functions.  
	}
	\label{fig:examples}
\end{figure}

Multiple sclerosis (MS) is a chronic, inflammatory demyelinating disease of the central nervous system in the brain.
Magnetic resonance imaging (MRI) can depict and characterize MS lesions for clinical diagnosis and assessment of disease progression.
These lesions are often highly heterogeneous with regards to appearance, location, size and shape (An example is shown in Fig.~\ref{fig:examples}).
Conventionally, lesions are segmented manually by a trained clinician, the process of which is tedious, time-consuming and has low reproducibility.
Many automated lesion segmentation algorithms have been developed to address this problem, but a clinically reliable technique is not yet available.

Unsupervised algorithms~\cite{schmidt2012automated, yu20143d} rely on carefully selected image features for segmentation and do not require training.
However, they are considered inferior to the more recently developed supervised algorithms, especially deep convolutional neural networks (CNNs)~\cite{zhang2019rsanet, zhang2020efficient, zhang2019multiple, aslani2019multi, hou2019cross}.
These deep learning models need to be trained, typically with region-based loss functions such as binary cross entropy (BCE) or Dice loss.
To tackle the data imbalance problem (only a small fraction of brain voxels belong to lesions), weighted BCE and Tversky loss~\cite{hashemi2018asymmetric} have been proposed. 



Geometric information has been utilized to further improve medical image segmentation~\cite{kervadec2018boundary, karimi2019reducing, xue2019shape, oda2018besnet, chen2016dcan}.
One approach is to develop an additional decoder architecture to generate geometric related feature maps for loss evaluations~\cite{xue2019shape, oda2018besnet, chen2016dcan}.
Another approach is based on shape- or boundary-aware loss function~\cite{kervadec2018boundary, karimi2019reducing} that performs geometric transformations on ground-truth or predicted probability map.
The distance transformation mapping (DTM) is used in both boundary (BD) loss~\cite{kervadec2018boundary} and Hausdorff distance (HD) loss~\cite{karimi2019reducing}, where each voxel in the transformation map presents the distance between it and the closest boundary of region-of-interests (ROIs).
The boundary-aware loss functions enforce networks to focus on the surface of the lesion ROIs, thereby addressing the large imbalance between lesion and background voxels.
While these loss functions can perform well on the segmentation of large objects, their performance is often not satisfactory for small MS lesions.

\begin{figure*}[!th]
	\centering
	\subfloat[FOG S]{\includegraphics[height=.28\columnwidth]{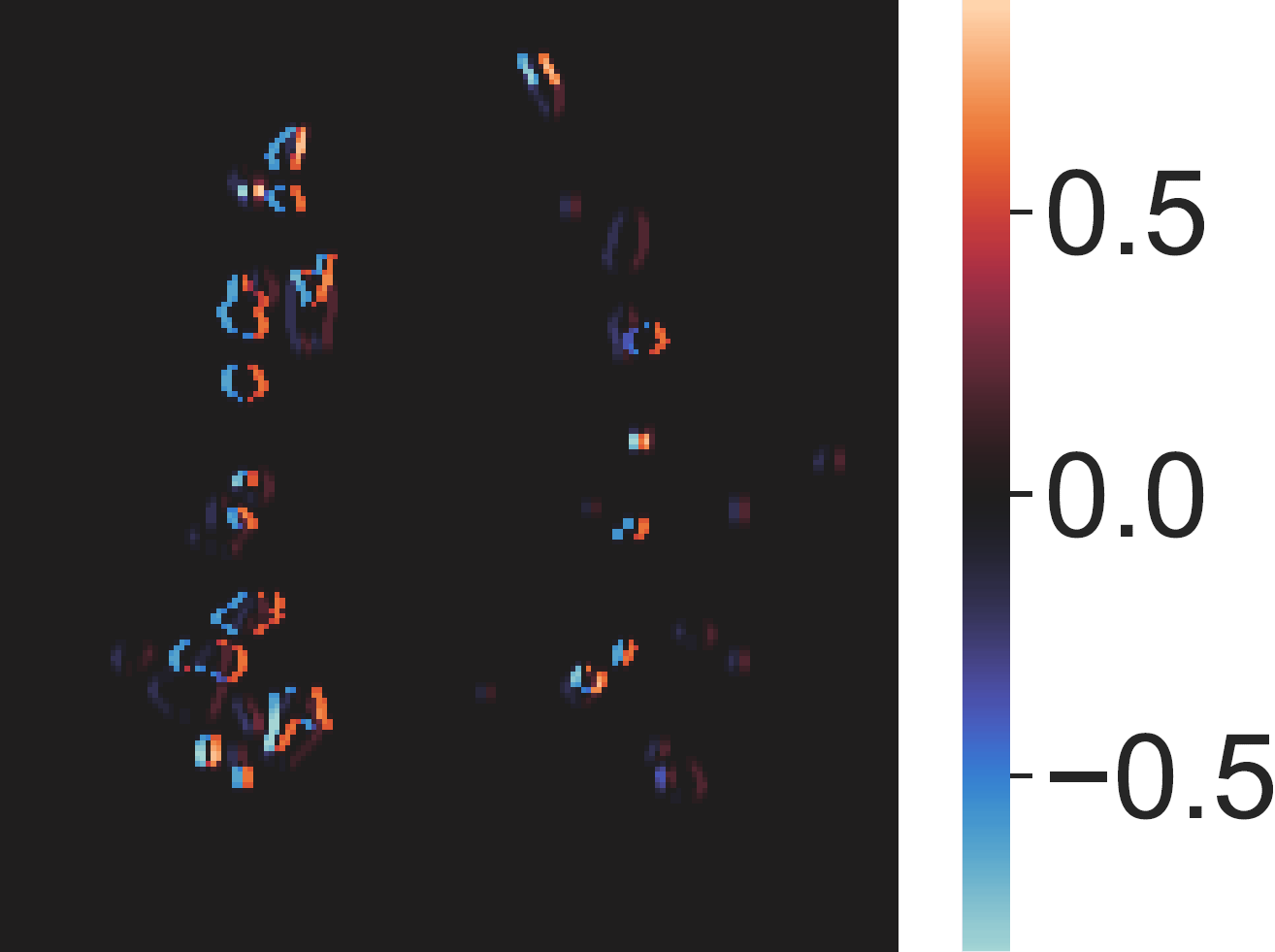} \label{fig:fog_s}}
	\hspace{0.5ex}
	\subfloat[FOG C]{\includegraphics[height=.28\columnwidth]{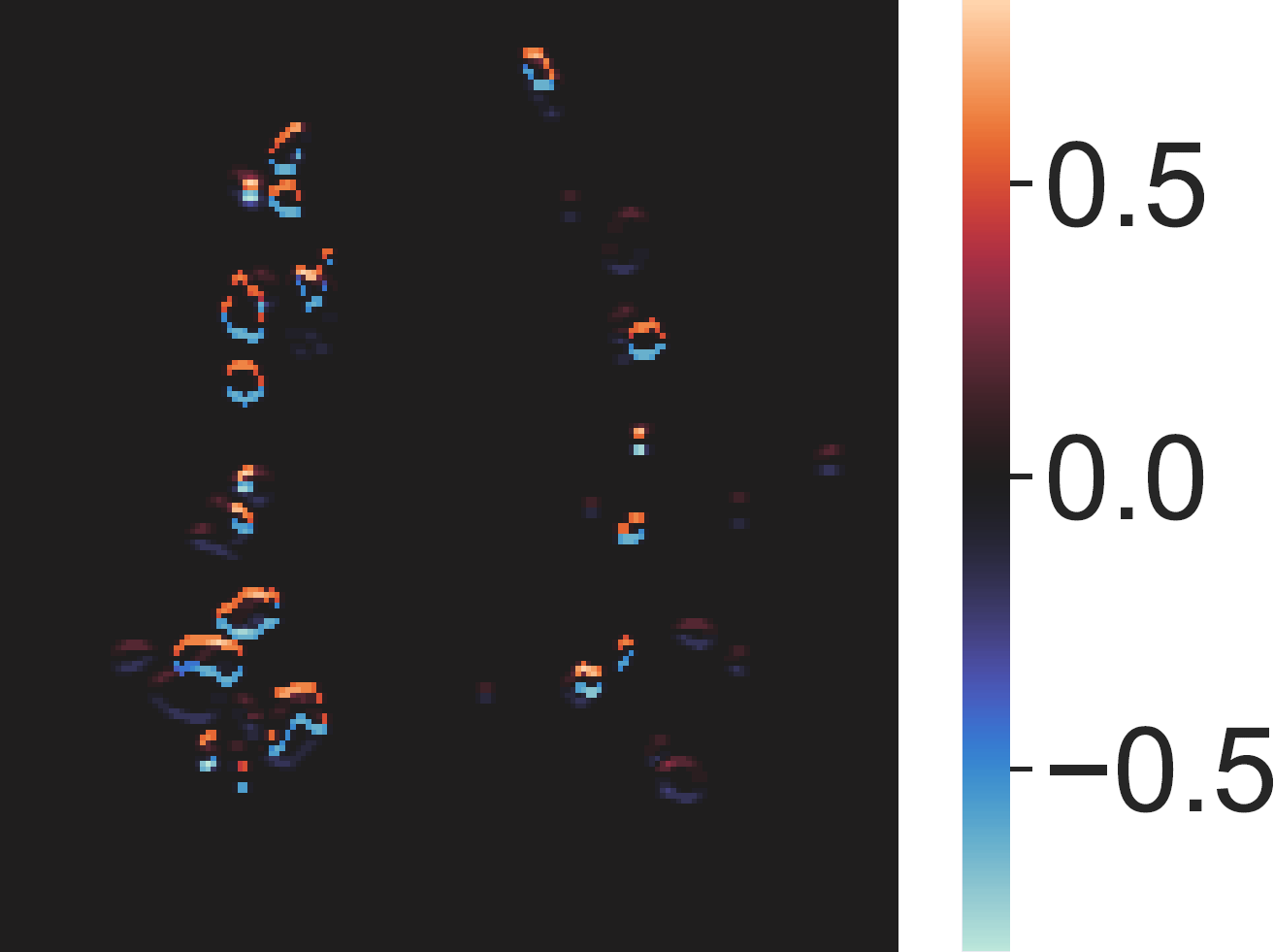} \label{fig:fog_c}}
	\hspace{0.5ex}
	\subfloat[FOG A]{\includegraphics[height=.28\columnwidth]{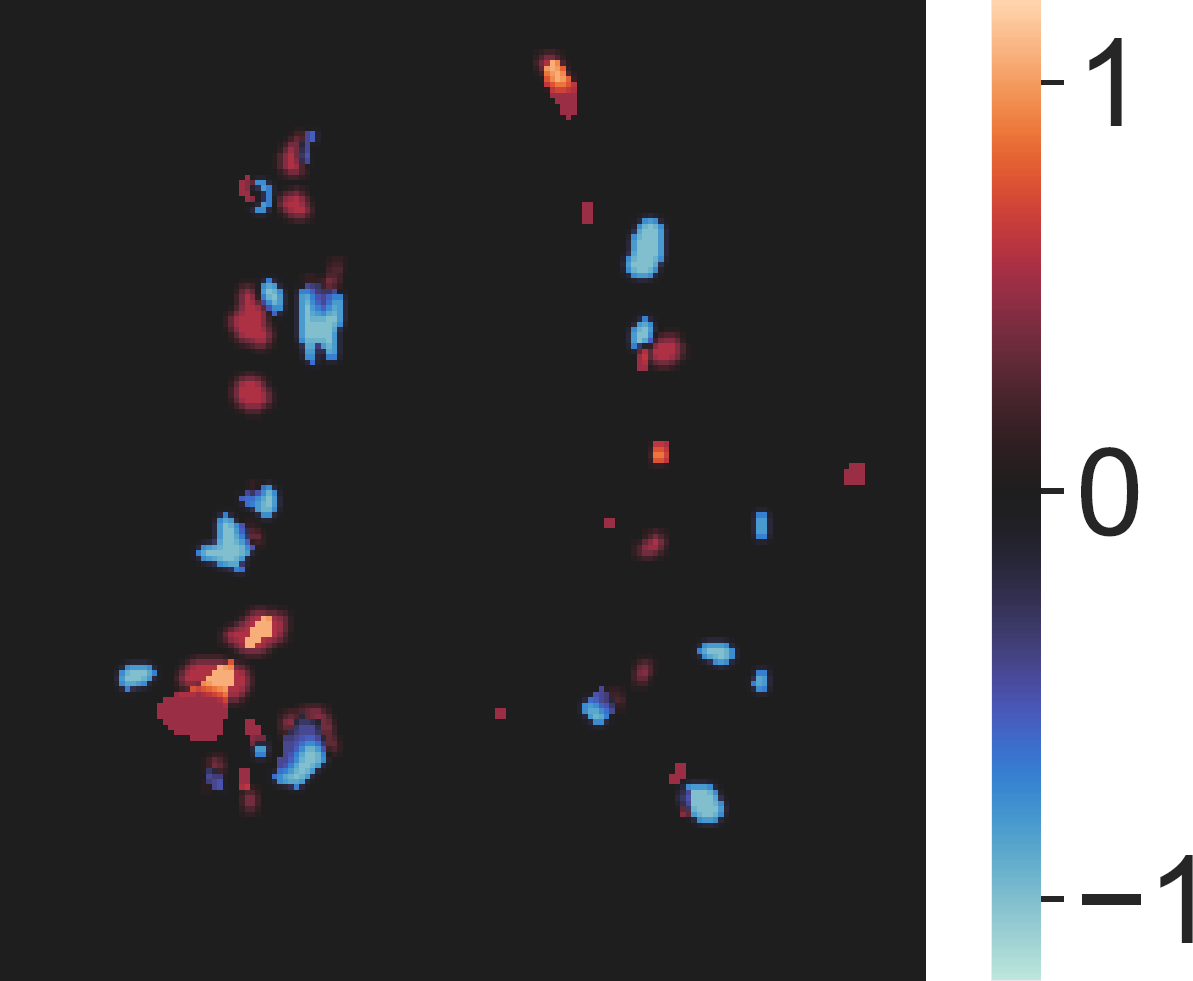} \label{fig:fog_a}}
    \hspace{0.5ex}
    \subfloat[SOG]{\includegraphics[height=.28\columnwidth]{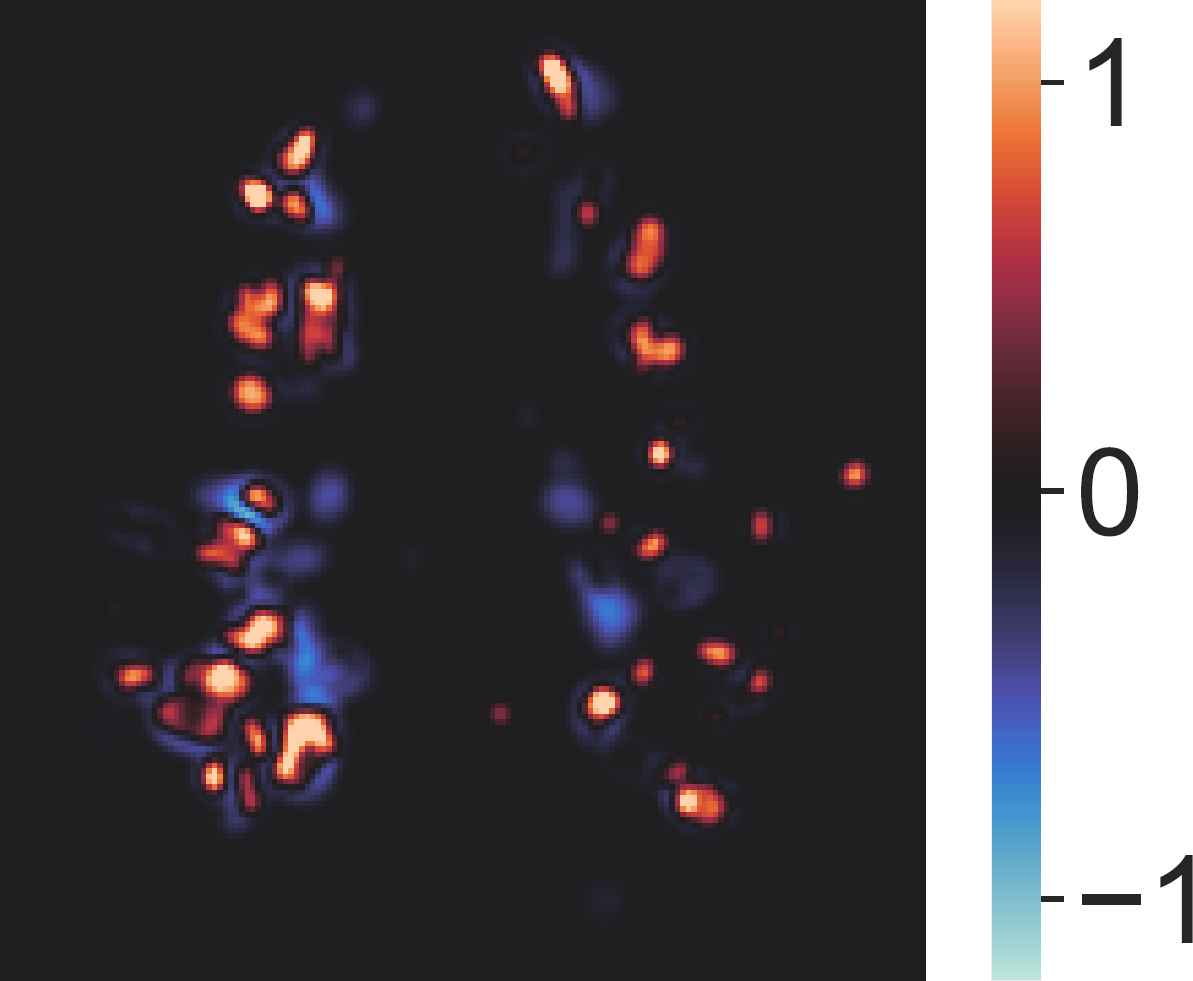} \label{fig:sog}}
	\hspace{0.5ex}
	\subfloat[DTM]{\includegraphics[height=.28\columnwidth]{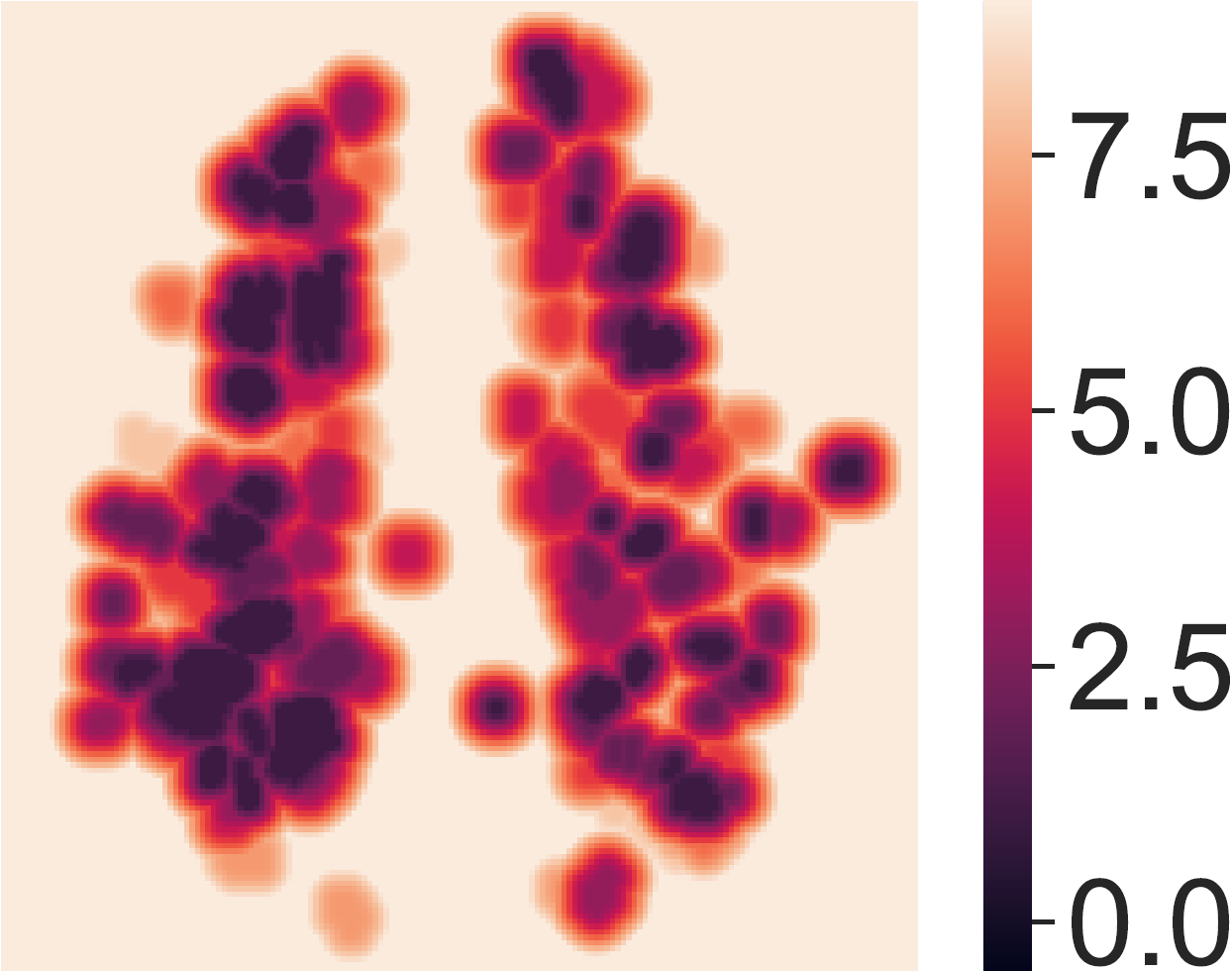} \label{fig:dtm}}
	\caption{ 
	    Example of geometric transformations applied to the ground-truth lesion mask shown in Fig.~\ref{fig:examples}.
	    FOG and SOG operators capture lesion edges, while DTM encodes the information of distance to edge.
	    Please note that all figures are slices from computation results in 3D space.
	}
	\label{fig:filter}
\vspace{-1ex}
\end{figure*}

In this paper, we propose a generalized \emph{geometric Loss} (GEO loss) formula for MS lesion segmentation.
Our method allows flexible and computationally inexpensive integration of region-based and geometric transformation based information in the design of CNN loss functions.
As an example, we introduced two new GEO loss functions based on lesion edge information and showed that the proposed method can outperform state-of-the-art algorithms.

\section{Methodology}

In this section, we will first describe the general form of GEO loss function and its relationships with traditional region-based and boundary-aware loss functions.
We will then derive two specific GEO loss functions based on edge information to improve lesion segmentation.


\subsection{Geometric Loss Formulation}

The proposed GEO loss combines volumetric and geometric correlations in a single module and has the following form:
\begin{equation}
    \mathcal{L}_{G} = \frac{\sum_{v \in \Omega}  \Theta(s_v, g_v)\Psi(s, g, v, \phi)}{\sum_{v \in \Omega}\Gamma (s_v, g_v)}
    \label{eqn:geo_loss_disc}
\end{equation}
Here $\Omega \subset \mathbb{R}^{3}$ is the spatial domain of an input 3D image, $v=(v_x, v_y, v_z)$ is the spatial position vector, $s$ is the output probability map, $s_v \in [0,1]$ is the value of $s$ at position $v$, $g$ is the ground-truth binary lesion mask, and $g_v \in \{0,1\}$ is the value of $g$ at position $v$.
The function $\Theta$ measures the voxel-wise volumetric correlations between the ground-truth lesion mask and the output probability map. 
$\phi$ denotes a spatially invariant operator defined on the spatial domain $\Omega$ which captures local geometric information such as edges and distance from edges.
Given $\phi$, the function $\Psi$ computes the voxel-wise geometric correlations between the ground-truth and the output map. 
$\Gamma$ serves as the overall normalization factor.



Unlike previous works~\cite{xue2019shape, oda2018besnet, chen2016dcan} that require auxiliary decoder networks or new network architectures, our GEO loss formula can be used in existing CNN models and allows flexible selection of loss functions for training.
The widely used region-based BCE and Dice loss functions can be seen as special cases of Eq.~\eqref{eqn:geo_loss_disc} with $\Psi=1$, and they can be derived by setting $\Theta(s_v, g_v) = -(g_v\log(s_v)+(1-g_v)\log(1-s_v))$, $\sum_{v \in \Omega}\Gamma (s_v, g_v) = 1$
, and $\Theta(s_v, g_v) = s_v+g_v - 2s_v g_v$ and $\Gamma (s_v, g_v) = s_v+g_v$, respectively.
Boundary-aware loss functions such as BD and HD can be derived from Eq.~\eqref{eqn:geo_loss_disc} with $\Gamma = 1$ by setting $\Psi(s, g, v, \phi) = \phi_D(g, v)$, $\Theta(s_v, g_v) = s_v$, and $\Psi(s, g, v, \phi) = \phi_D(g, v)^2 + \phi_D(s, v)^2$ and $\Theta(s_v, g_v) = (s_v - g_v)^2$ respectively.
The function $\phi_D(g, v)$ computes the distance between position $v$ and its closest ROI boundary of $g$.

\subsection{Geometric Loss Instantiations}


Since the boundary area of a 3D object is one order of magnitude smaller than its volume, loss computation in the boundary space can mitigate the large imbalance between lesion and background voxels and in turn benefits the segmentation of small lesions.
BD~\cite{kervadec2018boundary} and HD loss~\cite{karimi2019reducing} functions require expensive computation of DTM as a measure of lesion geometry, and misclassified voxels are re-weighted accordingly.
This re-weighting scheme is beneficial to segmenting large objects but merely contributes and can even be harmful to segmenting small lesions, because it puts less weights to misclassified small lesions.
Here we propose to apply computationally efficient convolutional filters such as first- and second-order gradient operators for edge enhancement in the loss functions.

\subsubsection{First Order Gradient (FOG) Loss.} 
Let the output of $\phi$ be a three-element vector, where $\phi(s, v)= (\frac{\partial s}{\partial v_x}, \frac{\partial s}{\partial v_y}, \frac{\partial s}{\partial v_z})$, $\phi(g, v)=(\frac{\partial g}{\partial v_x}, \frac{\partial g}{\partial v_y}, \frac{\partial g}{\partial v_z})$; also, letting $\forall~v \in \Omega$, $\Theta(s_v, g_v)=1$, and $\Psi(s, g, v, \phi) = ||\phi(s, v) - \phi(g, v)||^2$, we can define the FOG loss as following:
\begin{equation}
    \mathcal{L}_{F} = \frac{1}{|\Omega|} \sum_{v \in \Omega}  ||\phi(s, v) - \phi(g, v)||^2.
    \label{eqn:sobel_loss}
\end{equation}
We notice that when tracing a specific lesion, neuroradiologists usually examines surrounding slices on all axial, saggtial and coronal planes.
Based on the observation, we further design three variants of FOG loss that compute gradients on only one of the orthogonal planes, where  $\phi(s, v)=\frac{\partial s}{\partial v_i}$, $\phi(g, v)=\frac{\partial g}{\partial v_i}$, and $i$ enumerates $\{x,y,z\}$.

\subsubsection{Second Order Gradient (SOG) Loss.} 
The SOG loss is defined as the second-order differential operator which is the divergence of the gradient.
Based on the boundary property of SOG, letting $\phi(s, v)=\frac{\partial^2 s}{\partial^2 v_x} + \frac{\partial^2 s}{\partial^2 v_y} + \frac{\partial^2 s}{\partial^2 v_z}$, $\phi(g, v)=\frac{\partial^2 g}{\partial^2 v_x} + \frac{\partial^2 g}{\partial^2 v_y} + \frac{\partial^2 g}{\partial^2 v_z}$, and $\Theta(s_v, g_v) = |s_v - g_v|$, SOG loss is derived as:
\begin{align}
    \mathcal{L}_{S} & = \frac{1}{|\Omega|} \sum_{v \in \Omega} |s_v-g_v|\phi(g, v)
    \label{eqn:lap_loss}
\end{align} 
Eq.~\ref{eqn:lap_loss} is the one-sided SOG loss, and the two sided SOG loss can be obtained by replacing the term $\phi(g,v)$ with $(\phi(g,v) + \phi(s,v))$.
Fig.~\ref{fig:filter} shows an example of the proposed geometric transformation based on first- and second-order graident in comparison with the DTM transformation.
\section{Experimental Results}

\begin{figure}[!b]
	\centering
	\subfloat{\includegraphics[width=.5\columnwidth]{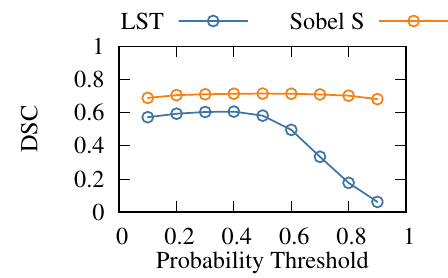}  \label{fig:dsc}}
	\subfloat{\includegraphics[width=.5\columnwidth]{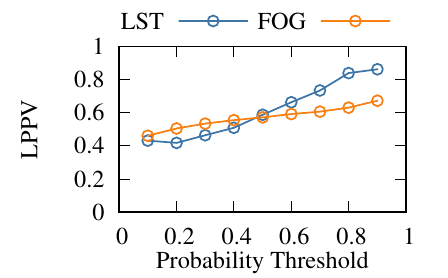} \label{fig:ppv}} \\
	\subfloat{\includegraphics[width=.5\columnwidth]{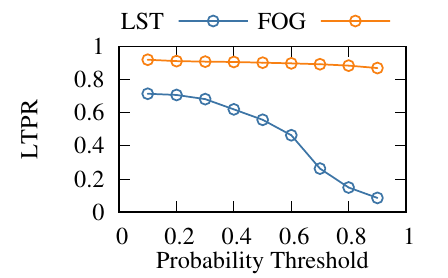} \label{fig:ltpr}}
	\subfloat{\includegraphics[width=.5\columnwidth]{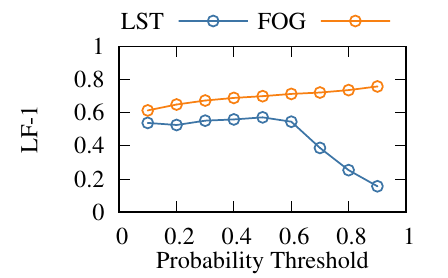} \label{fig:lf_1}}
	\caption{ 
	    Performance comparison between LST and FOG (co-trained with Dice loss) on DSC, LPPV, LTPR, and LF-1. 
	    Nine probability thresholds from $0.1$ to $0.9$ are applied.
	}
	\label{fig:lga}

\end{figure}
In this section, we compared the proposed GEO loss functions with state-of-the-art boundary-aware loss functions.
Two datasets with different scales were used for performance evaluation.
One small-scale dataset (GE-30) consisted of co-registered T1, T2, and T2-FLAIR images ($0.7mm \times 0.7mm \times 3mm$ voxel size) acquired at a 3T GE scanner from 30 MS patients.
Another large-scale dataset (SI-170) consisted of co-registered T1, T2, and T2-FLAIR images ($1mm \times 1mm \times 1mm$ voxel size) acquired at a 3T SIEMENs scanner from 170 MS patients.
The ground-truth masks of both datasets were traced by a neuroradiologist with 8 years of experience. 
The small-scale GE-30 was used to show that our methods are more robust for training deep neural networks with limited training samples.
The large-scale SI-170 was used to demonstrate that our methods are also efficient in dealing with lesion segmentation with complicated contextual details.

\subsection{Implementation Details}
We used PyTorch~\cite{paszke2019pytorch} to implement all loss functions as well as our backbone CNN architecture 3D U-Net~\cite{cciccek20163d}. 
All experiments were performed on a computer with a Nvidia Titan Xp GPU.
For GE-30 dataset, 15, 5, and 10 subjects were used for training, validation, and testing, respectively.
For GE-170 dataset, 119, 17, 34 subjects were used for training, validation, and testing, respectively.
All images were linearly co-registered using FSL~\cite{smith2004advances} FLIRT command, followed by image intensity normalization.
We further used random crop, intensity shifting, and elastic deformation for data augmentation.
To train each model, we adopted Adam \cite{kingma2014adam} optimizer with an initial learning rate of $1\mathrm{e}{-3}$ (weight decay of $1\mathrm{e}{-6}$), and the batch size was four. 
The learning rate was halved at $50\%$, $70\%$ and $90\%$ of the total training epochs ($100$) for optimal convergence. 

We used Dice similarity coefficient (DSC), lesion-wise true positive rate (LTPR), lesion-wise positive predictive value (LPPV), and lesion-wise F1 score (L-F1) as evaluation metrics.
LTPR, LPPV, and L-F1 are defined as $\text{LTPR} = \frac{\text{TPR}}{\text{GL}}$, $\text{LPPV} = \frac{\text{TPR}}{\text{PL}}$, and $\text{L-F1}=2\frac{\text{LTPR}\cdot \text{LPPV}}{\text{LTPR}+\text{LPPV}}$, where $\text{TPR}$ denotes the number of lesions in the ground-truth segmentation that overlap with a lesion in the produced segmentation, and $\text{GL}$ and $\text{PL}$ are the number of lesions in the ground-truth segmentation and the predicted segmentations, respectively.
DSC quantifies the voxel-wise overlap between the output and the ground-truth, while, LTPR, LPPV and L-F1 measure the lesion-wise detection accuracy.

\begin{table}[!t]
\caption{Ablation study on the GE-30 dataset.}
\begin{center}
    \resizebox{0.75\columnwidth}{!}
    {
    \begin{small}
    \begin{tabular}{l|cccc}
    \hline
    \hline
                Methods                                      &DSC    &LPPV   &LTPR   &L-F1  \\
    \hline
                Dice                                         &0.705  &0.580  &0.834  &0.684 \\
                Dice + FOG                             &\underline{0.712}  &\textbf{0.627}  &0.863  &\textbf{0.726} \\ 
                Dice + FOG S                           &\textbf{0.715}  &0.592  &\textbf{0.898}  &0.714 \\
                Dice + FOG C                           &0.708  &0.602  &\underline{0.889}  &\underline{0.718} \\
                Dice + FOG A                           &0.702  &0.586  &0.862  &0.698 \\ 
                Dice + SOG One                           &0.708  &\underline{0.606}  &0.857  &0.710 \\
                Dice + SOG Two                           &0.706  &0.600  &0.855  &0.705 \\
    \hline
    \hline
    \end{tabular}
    \end{small}
    \label{tab:ablation_study}
}
\end{center}
\vskip -0.15in
\end{table}

\subsection{Effectiveness of the Geometric Loss Functions}

In this section, we present our ablation study to show the effectiveness of variants of the proposed loss functions and present comparison with other state-of-the-art boundary-aware loss functions.
``FOG" denotes the loss function defined in Eq.~\eqref{eqn:sobel_loss}, and "FOG S",  "FOG C" and  "FOG A" represent FOG loss in the saggital, coronal and axial directions respectively.
``SOG One" and ``SOG Two" indicate one-sided and two-sided SOG loss in Eqn.~\eqref{eqn:lap_loss}.
``HD" and ``BD" are loss functions adopted from the previous literature~\cite{kervadec2018boundary, karimi2019reducing}.
All geometric based loss functions are applied together with region based Dice loss to get the optimal performance. 
``LST" is a well-known tool~\cite{schmidt2012automated} for MS lesion segmentation, where unsupervised lesion growth algorithm is applied.
For all comparing methods, we employed their open-source implementations and optimized their performance for MS lesion segmentation.
In following three tables, a bold number in the table indicates the best score in its column, and an underlined number is the second best score in its column.

\subsubsection{Ablation study}

We used the GE-30 dataset to conduct the ablation study, and Table~\ref{tab:ablation_study} summarizes the performance of variants of the proposed GEO loss functions.
We can see from Table~\ref{tab:ablation_study} that all variants of the proposed GEO loss functions outperformed region-based Dice loss in all evaluation metrics.
FOG S achieved the best DSC and LTPR scores, while FOG C performed similar to FOG C.
However, interestingly, FOG A was not as good as its counter parts.
Besides the shape information, boundaries along sagittal and coronal directions can provide additional lesion location information (corresponds to our clinical observation), while the boundary along the axial direction encodes similar information as region based loss.
FOG combines region gradient information from three orthogonal directions and achieves the best LPPV and L-F1 scores among all.
One sided SOG achieved slightly better performance than two sided SOG.
Based on the performance of all these variants, FOG and SOG one are picked up to compare with other state-of-the-art boundary-aware methods.

\subsubsection{Results of Unsupervised Methods.}

We used the GE-30 dataset to demonstrate the effectiveness of unsupervised methods.
As Fig.~\ref{fig:lga} shows, the supervised deep learning model outperformed the unsupervised LST algorithm.
The deep learning model is not sensitive to the thresholding paramter in terms of DSC and LTPR, and thus it is easy to find the optimal trade-off between LTPR and LPPV.



\begin{table}[!t]
\caption{Performance comparison on the GE-30 dataset.}

\begin{center}
    \resizebox{0.75\columnwidth}{!}
    {
        \begin{small}
        \begin{tabular}{l|cccc}
        \hline
        \hline
                    Methods                                  &DSC    &LPPV   &LTPR   &L-F1  \\
        \hline
                    LST \cite{schmidt2012automated}          &0.594  &0.418  &0.708  &0.526 \\
                    Dice                                     &0.705  &0.580  &0.834  &0.684 \\
                    Dice + HD \cite{karimi2019reducing}      &0.502  &0.393  &0.608  &0.477 \\      
                    Dice + BD \cite{kervadec2018boundary}    &0.704  &0.594  &\textbf{0.875}  &0.708 \\
                    Dice + FOG                               &\underline{0.712}  &\textbf{0.627}  &\underline{0.863}  &\textbf{0.726} \\ 
                    Dice + SOG One                           &0.708  &\underline{0.606}  &0.857  &\underline{0.710} \\
        \hline
        \hline
        \end{tabular}
        \end{small}
        \label{tab:ge_30}
    }
\end{center}
\vskip -0.15in
\end{table}

\subsubsection{Results of Supervised Methods.}

We compared our proposed GEO loss functions with other state-of-the-art boundary-aware loss functions as well as region based Dice loss on GE-30 and SI-170 datasets.
BD loss applied the scheduling to trade-off region based loss terms and boundary based loss terms as the literature~\cite{kervadec2018boundary} suggests.
All of our proposed loss as well as HD loss~\cite{karimi2019reducing} functions use $1.0$ as the GEO loss coefficient.

Table~\ref{tab:ge_30} reports detailed comparisons on the GE-30 dataset between traditional and our proposed loss functions.
Dice and HD loss functions fall behind by other loss functions with quiet a gap.
HD is originally designed for improving the Hausdorff distance metric; though it performs well on segmenting large objects~\cite{karimi2019reducing}, it fails at MS lesion segmentation, where dozens of lesions inside a single brain can vary greatly in terms of location and shape.
SOG one and BD have similar composition of the loss structure, where they both use the product of the region factor and the geometric factor to re-weight mis-classified voxels, but they used different geometric transformations, and second-order gradient operator is more effective than DTM in segmenting lesions.
BD has achieved slightly better LTPR than our proposed GEO loss functions.
However, our FOG outperformed BD in LPPV and L-F1 by $6\%$ and $3\%$ respectively.

Table~\ref{tab:si_170} reports detailed comparisons on the GE-170 dataset between traditional and our proposed loss functions.
Compared with the GE-30 dataset, SI-170 has more complicated image details, as the slice in the SI-170 is three times thinner than that in the GE-30.
Similarly, Dice and HD loss functions fall behind by other loss functions.
Interestingly, in this more challenging data, our proposed FOG and SOG One outperformed BD
in all four metrics with a significant margin.

In general, experiments on both large-scale dataset GE-30 and dataset SI-170 with limited training samples demonstrated the effectiveness and the robustness of our proposed GEO loss functions.
Effectiveness on GE-30 shows that our methods can generalize well even with limited training samples presented.
Results on SI-170 shows that our methods can capture complicated contextual details.

\begin{table}[!t]
\caption{Performance comparison on the SI-170 dataset.}

\begin{center}
    \resizebox{0.75\columnwidth}{!}
    {
    \begin{small}
    \begin{tabular}{l|cccc}
    \hline
    \hline
                Methods                                      &DSC    &LPPV   &LTPR   &L-F1  \\
    \hline
                LST \cite{schmidt2012automated}              &0.559  &0.526  &\underline{0.870} &0.656 \\
                Dice                                         &0.726  &0.598  &0.869  &0.725 \\
                Dice + HD \cite{karimi2019reducing}          &0.635  &0.618  &0.845  &0.714 \\
                Dice + BD \cite{kervadec2018boundary}        &0.725  &0.637  &0.855  &0.730 \\
                Dice + FOG                                   &\textbf{0.737}  &\textbf{0.666}  &\textbf{0.873}  &\textbf{0.756} \\
                Dice + SOG One                               &\underline{0.735}  &\underline{0.664}  &0.860  &\underline{0.749} \\
    \hline
    \hline
    \end{tabular}
    \end{small}
    \label{tab:si_170}
}
\end{center}
\vskip -0.15in
\end{table}
\section{Conclusions}

We presented a novel GEO loss formula to allow flexible and computationally efficient integration traditional region-based and boundary-aware loss functions.
Two new loss functions were derived based on the first- and second-order gradient operators to utilize lesion edge information.
These loss functions and their variants outperformed state-of-the-art methods and could achieve a good trade-off between LTPR and LPPV to improve the overall accuracy.  

\newpage

\bibliographystyle{IEEEbib}
\bibliography{strings,refs}

\end{document}